\documentclass[10pt,twocolumn,letterpaper]{article}

\usepackage{cvpr}              

\usepackage[dvipsnames]{xcolor}

\definecolor{cvprblue}{rgb}{0.21,0.49,0.74}
\usepackage[pagebackref,breaklinks,colorlinks,citecolor=cvprblue]{hyperref}

\def\paperID{***} 
\def\confName{CVPR}
\def\confYear{2024}

\usepackage{epsfig}
\usepackage{multicol}
\usepackage{multirow}
\usepackage[accsupp]{axessibility}

\title{PromptCIR: Blind Compressed Image Restoration with Prompt Learning}

\author{Bingchen Li\textsuperscript{\rm 1}, Xin Li\textsuperscript{\rm 1}, Yiting Lu\textsuperscript{\rm 1}, Ruoyu Feng\textsuperscript{\rm 1}, Mengxi Guo\textsuperscript{\rm 2}, Shijie Zhao\textsuperscript{\rm 2}, Li Zhang\textsuperscript{\rm 2}, Zhibo Chen\textsuperscript{\rm 1}\footnotemark[2]\\
\textsuperscript{\rm 1}University of Science and Technology of China \quad
\textsuperscript{\rm 2}Bytedance Inc. \\
\tt\small \{lbc31415926, lixin666, luyt31415, ustcfry\}@mail.ustc.edu.cn, \\ \tt\small \{guomengxi.qoelab, zhaoshijie.0526, lizhang.idm\}@bytedance.com, chenzhibo@ustc.edu.cn
}

\begin{document}
\maketitle

\begin{abstract}
Blind Compressed Image Restoration (CIR) has garnered significant attention due to its practical applications. It aims to mitigate compression artifacts caused by unknown quality factors, particularly with JPEG codecs. Existing works on blind CIR often seek assistance from a quality factor prediction network to facilitate their network to restore compressed images. However, the predicted numerical quality factor lacks spatial information, preventing network adaptability toward image contents. Recent studies in prompt-learning-based image restoration have showcased the potential of prompts to generalize across varied degradation types and degrees. This motivated us to design a prompt-learning-based compressed image restoration network, dubbed PromptCIR, which can effectively restore images from various compress levels. Specifically, PromptCIR exploits prompts to encode compression information implicitly, where prompts directly interact with soft weights generated from image features, thus providing dynamic content-aware and distortion-aware guidance for the restoration process. The light-weight prompts enable our method to adapt to different compression levels, while introducing minimal parameter overhead. Overall, PromptCIR leverages the powerful transformer-based backbone with the dynamic prompt module to proficiently handle blind CIR tasks, winning first place in the NTIRE 2024 challenge of blind compressed image enhancement track. Extensive experiments have validated the effectiveness of our proposed PromptCIR. The code is available at \url{https://github.com/lbc12345/PromptCIR-NTIRE24}.

\end{abstract}

\section{Introduction}
The rapid growth of digital images and videos has necessitated the utilization of lossy compression techniques~\cite{JPEG,HEVC,bross2021overviewVVC,wu2021learned,li2021task} to optimize storage and bandwidth utilization. Among popular compression codecs, JPEG is favored for its computational efficiency and straightforward implementation. It compresses an image by segmenting images into $8 \times 8$ blocks, applying discrete cosine transform (DCT), and quantizing the coefficients. During quantizing, the information is inevitably lost, resulting in unfavorable compression artifacts. The degree of compression artifacts is affected by the quality factor ranging from 0 to 100, with lower values indicating worse image qualities.

With the vast development of deep neural networks (DNNs), researchers have explored the application of DNNs in image restoration tasks~\cite{liang2021swinir,conde2022swin2sr,wang2022uformer,li2020learning,li2023learning,li2023diffusion,liu2020lira,pang2020fan}. Pioneering works~\cite{arcnn} utilizes convolutional neural networks (CNNs) to tackle compressed image restoration (CIR) problems. In recent years, Transformers~\cite{ViT,liu2021swin,liu2022swin} have showcased powerful representative capabilities with self-attention modules. Models incorporating transformer blocks~\cite{liang2021swinir,wang2022uformer,zamir2022restormer,conde2022swin2sr,li2022hst} have surpassed previous CNN-based methods, restoring more rich texture details. However, most methods focus on restoring compressed images with predefined JPEG quality factors, significantly restricting their practicality in real-world scenarios where the quality factors are often unknown. To mitigate this problem, a series of works~\cite{fbcnn,blindqf,blindqf2,blindqf3} attempt to predict the quality factor as an auxiliary guidance, avoiding predefining the quality factor before network training.

Although straightforward, the prediction network still faces two challenges: 1) The predicted numerical quality factor loses all the spatial-wise information, lacking content-aware adaptability which is essential for image restoration (IR) tasks. 2) The prediction network usually causes significant parameter overhead, which constrains its deployment on edge devices like mobile phones~\cite{blindcontrastive}. To tackle the above challenges, we propose PromptCIR in this paper, a prompt-learning-based blind CIR approach.

Recent advances in prompt-learning-based IR~\cite{promptir,ma2023prores,pip,daclip} have demonstrated the great generalization ability of prompts for universal IR tasks. Particularly, PromptIR~\cite{promptir} leverages lightweight prompts to encode degradation information as a guidance for restoration networks. Inspired by this, we explore an efficient method of restoring blind compressed images with prompts, dubbed PromptCIR. Different from previous prediction-based methods which directly estimate numerical quality factors for guidance, PromptCIR learns to encode compression information implicitly through the interaction between prompt components, and prompt weights generated from distorted image features, thus providing more content-aware adaptability for blind restorations with a few additional parameters. However, it is noteworthy that most existing prompt-learning-based IR methods~\cite{promptir,ma2023prores} set the prompt size to be the same as the size of image feature, constraining the generalization ability for various input sizes. To mitigate this problem, UCIP~\cite{li2024ucip} employs a content-aware dynamic prompt. Specifically, they set the prompt size to $1 \times 1$ and generate spatial-wise prompt weights with the same resolution of image features to avoid the potential problem of overfitting to training image size. Additionally, to improve the content-aware and distortion-aware representation ability, they use several prompt bases to dynamically encode task-adaptive information. Inspired by this, we adopt the same prompts design from UCIP~\cite{li2024ucip} to adaptively encode compression quality information in PromptCIR, while preserving necessary spatial-wise knowledge. 

Concretely, the U-shape structure has demonstrated strong representation ability for IR tasks~\cite{ushapeIR,ushapeIR2,ushapeIR3,ushapeIR4,wang2022uformer,zamir2022restormer,promptir}. Building on the latest powerful restoration model Restormer~\cite{zamir2022restormer}, we design our PromptCIR with a U-shape structure to hierarchically restore image features. Moreover, recent studies~\cite{li2023lsdir,yang2023hq50k,supir} have showcased that larger datasets can further boost the performance of restoration networks in terms of both objective and subjective qualities, as they encompass wider-range of texture details and more fine-grained semantics. Following their success, we further enhance the performance of our method by utilizing the open-source high-quality dataset LSDIR~\cite{li2023lsdir}. With dynamic prompts, representative network structure, and a powerful dataset, our PromptCIR has achieved first place in the NTIRE 2024 challenge of blind compressed image enhancement track~\cite{yang2024ntire}.

\section{Related Works}

\subsection{Compressed Image Restoration}
Compressed Image Restoration aims to reduce compression artifacts caused by various codecs, especially for JPEG artifact reduction. Early works mainly focus on designing filters in the DCT domain. ARCNN~\cite{arcnn}, as a pioneer work, first introduced a deep-learning-based convolutional neural network (CNN) to restore compressed images. To further improve the representation ability of the network, some works explore non-local attention~\cite{rnan} and residual-learning~\cite{rdn}, which achieve great restored results. Apart from the above studies in the pixel domain, DDCN~\cite{DDCN}, DMCNN~\cite{DMCNN} and $\text{D}^3$~\cite{wang2016d3} utilize DCT domain priors to further facilitate the restoration of compressed images. Inspired by these, Fu \etal~\cite{Fu_2019_ICCV} leverages dual domain knowledge with sparse coding to build a more compact and explainable restoration network. Similarly,~\cite{zheng2019implicit,blindqf2} make the most of dual domain information to achieve favorable restoration results. Recently, transformer-based methods have demonstrated great potential in IR tasks~\cite{liang2021swinir,wang2022uformer,zamir2022restormer,hat}. Particularly, shifted window-based attention further boost the performance of CIR~\cite{liang2021swinir} and super-resolution~\cite{li2023cswin2sr,li2022hst,conde2022swin2sr} by long-range pixel information interaction.

However, most of the aforementioned methods rely on a predefined compression quality factor, limiting their usage in practical scenarios~\cite{fbcnn}. Ehrlich \etal~\cite{QGAC} utilizes the quantization table as prior knowledge to facilitate the restoration of blind compressed images. Similarly, some works~\cite{blindqf,blindqf2,blindqf3} propose to estimate the compression quality factor to handle blind CIR problems. Particularly, FBCNN~\cite{fbcnn} proposes a quality factor predictor to estimate the quality factor from image features and a controller to guide the subsequent restoration network, achieving state-of-the-art performance on blind CIR. Nevertheless, the predicted numerical quality factor lacks spatial information on image features, restricting their content-aware adaptability. Moreover, such a method requires a carefully designed predictor and controller network with non-negligible overhead in terms of additional parameters. Different from explicitly predicting quality factor, we adopt lightweight prompts to implicitly encode compression information while preserving spatial-wise flexibility.

\subsection{Prompt-Learning-Based Image Restoration}

\begin{figure*}[ht]
    \centering
    \includegraphics[width=1.0\linewidth]{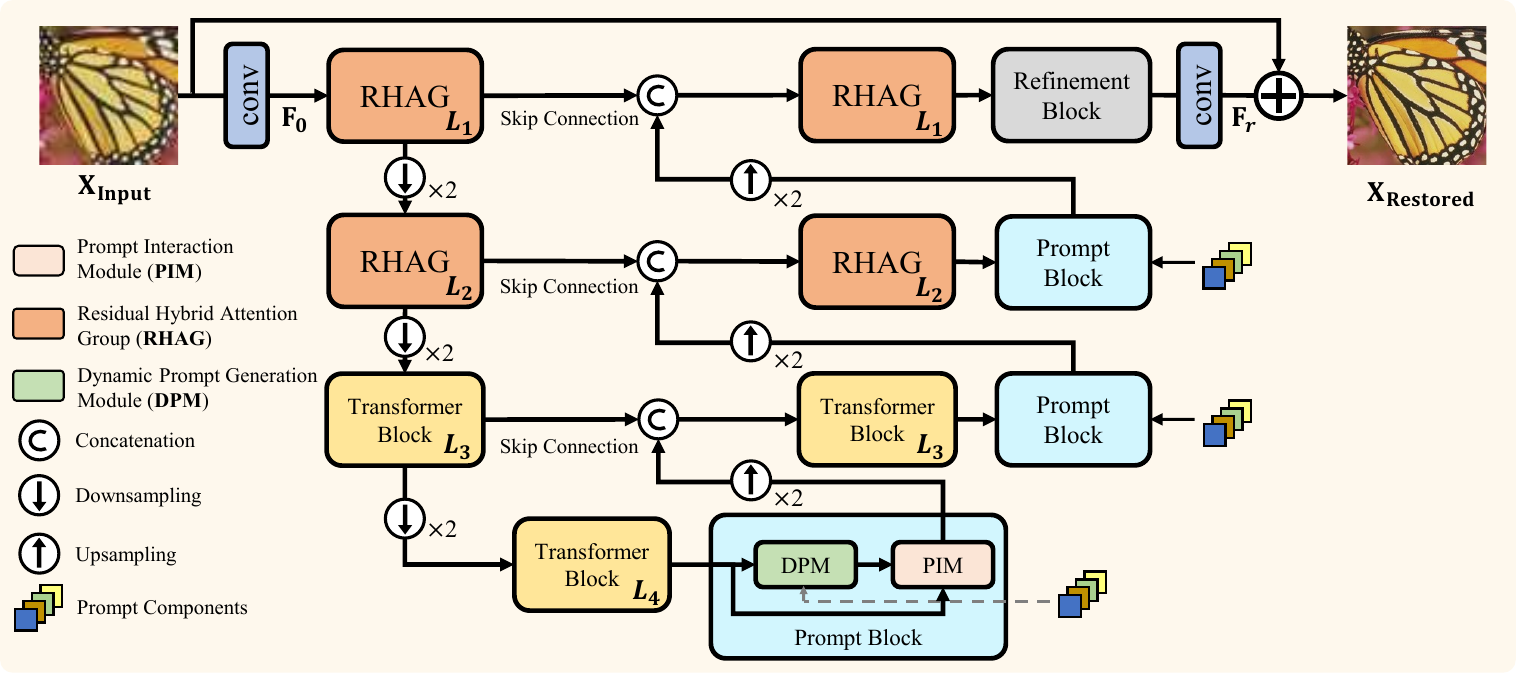}
    \caption{The overall framework of our proposed PromptCIR. We introduce prompt learning for blind compressed image restoration. To preserve content-aware knowledge while efficiently encoding distortion-aware information, we utilize DPM from UCIP~\cite{li2024ucip} to provide implicit guidance for the restoration process. To enhance the representation ability of the network, we replace the transformer blocks in the first two stages with RHAG inherited from HAT~\cite{hat}.}
    \label{fig:framework}
\end{figure*}

With the exponential growth of language and vision model~\cite{GPT3,openai2023gpt4,touvron2023llama,liu2024visual,zheng2023judging}, prompt learning has begun to play an increasingly important role~\cite{visual_prompt_tuning,visual_prompt_tuning2}. Inspired by the success of prompt learning in language and vision tasks, several studies have started to employ prompts in IR tasks~\cite{promptir,ma2023prores,daclip,pip}. Early work PromptIR~\cite{promptir} enhances powerful Restormer~\cite{zamir2022restormer} backbone, applying a set of prompts to encode distortion information and identify different restoration tasks such as image denoising, dehazing, and deraining. The prompts will interact with weights generated from image features to perform implicit guidance for the restoration network. Simultaneously, ProRes~\cite{ma2023prores} leverages an image-like prompt during fine-tuning to conduct efficient prompt-tuning for restoration. Moreover, PIP~\cite{pip} employs a dual prompt approach with one prompt responsible for distortion types and another prompt for texture enhancement. Prompt-learning-based IR also attracts interests from diffusion-based models. DA-CLIP~\cite{daclip} uses prompts during diffusion steps to facilitate universal IR, while PromptRR~\cite{wang2024promptrr} utilizes diffusion to refine prompts for better reflection removal. In this work, we explore prompts with blind CIR task. We use a set of dynamic prompts~\cite{li2024ucip} to perform comprehension of distortion levels while maintaining the capability of learning content-aware information for blind restoration.

\section{Method}

In this section, we first demonstrate the adaptability of our prompts towards content-aware and distortion-aware modeling. Then, we introduce the advantage of hybrid attention block~\cite{hat} against the original transformer block used in Restormer~\cite{zamir2022restormer}. Finally, we present the overall framework of our proposed PromptCIR.

\subsection{Prompt Block}
\label{sec:dpm}
\noindent\textbf{Preliminary.} With the development of large models~\cite{openai2023gpt4,touvron2023llama,liu2024visual}, prompt learning has played essential role for efficient tuning. In such a training scheme, the original large model is usually fixed~\cite{daclip,li2024sed}, while the lightweight prompts are optimized for new tasks. However, in IR fields, the well-trained large foundation model for various degradation restoration is missing. To restore images with unknown degradations, researchers use prompts as an efficient component~\cite{ai2023multimodal} to adaptively encode distortion-aware information to guide the restoration process with a single all-in-one restoration model, while the prompts and the model are jointly \textit{trained from-scratch}~\cite{promptir,pip,li2024ucip}. Blind CIR is very similar to all-in-one restoration. In blind CIR, the distortion type is consistent, but its degree remains unknown to the model. Therefore, using prompts in blind CIR would be more effective in modeling distortion-aware information.

\begin{figure*}[ht]
    \centering
    \includegraphics[width=1.0\linewidth]{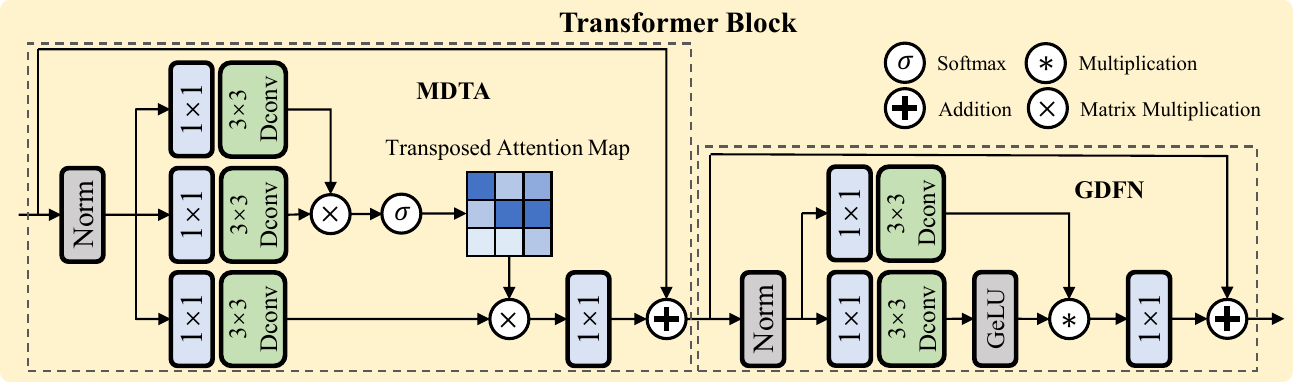}
    \caption{The structure of transformer block~\cite{zamir2022restormer,promptir}. It is composed of two modules, including the Multi Dconv head transposed attention module (MDTA) and the Gated Dconv feed-forward network (GDFN). Compared to traditional self-attention block, transposed attention mechanism provides more efficient information extraction with less computational complexity.}
    \label{fig:transformer}
\end{figure*}

\begin{figure}[h!]
    \centering
    \includegraphics[width=1.0\linewidth]{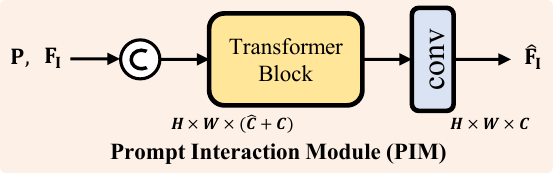}
    \caption{The structure of PIM. Here a $1\times 1$ convolution is used to reduce channel numbers to the same as input features.}
    \label{fig:PIM}
\end{figure}

\noindent\textbf{Structure.} The prompt block is composed of a dynamic prompt generation module (DPM) and a prompt interaction module (PIM). Compared to other prompt generation module, DPM~\cite{li2024ucip} features more adaptability in content-aware representation, as it uses several prompt bases to model spatial information. Taking prompt components with $N$ prompt bases $P_b \in \mathbb{R}^{1 \times 1 \times \hat{C}}$ and image features $F_I \in \mathbb{R}^{H \times W \times C}$ as inputs, DPM first generates weights $w_i \in \mathbb{R}^{H \times W \times N}$ directly from $F_I$, thus preserving distortion-aware adaptability. Next, the weights interact with the prompt bases in a manner similar to the operation of attention mechanisms to form the final prompts $P \in \mathbb{R}^{H \times W \times \hat{C}}$. Each spatial position of $P$ is dynamically composed by prompt bases through weights, thus realizing content-aware flexibility, which is crucial for blind CIR. This process can be formulated as:
\begin{equation}
    P = \texttt{Conv}\left( \sum_{b=1}^{N} w_i \odot P_b \right),\quad w_i = \texttt{Softmax}(\texttt{MLP}(F_I))
\end{equation}

As PIM is not our contribution, we directly adopt the well-designed module from PromptIR~\cite{promptir}. PIM (shown in Fig.~\ref{fig:PIM}) leverages a transposed self-attention transformer block to efficiently fuse information of prompt guidance with image features, as shown in Fig.~\ref{fig:transformer}.

\subsection{Hybrid Attention Block}
\label{sec:hab}

\begin{figure}
    \centering
    \includegraphics[width=1.0\linewidth]{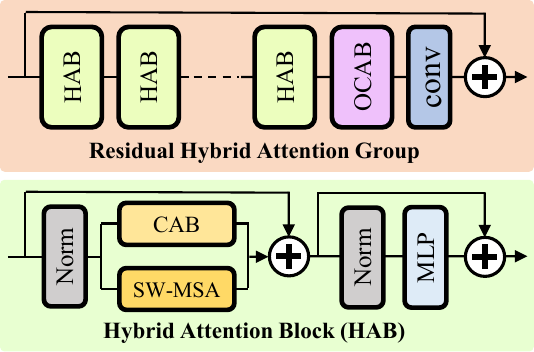}
    \caption{(Upper) illustrates the structure of RHAG~\cite{hat}, it combines several hybrid attention blocks (Bottom) with an overlapped cross-attention block (OCAB) followed by one convolution. Such design enhances the representation capabilities of the network in both local and global information extraction, and highly meets our needs for blind CIR.}
    \label{fig:RHAG}
\end{figure}

Blind CIR task requires that the network possess strong representation capability, encompassing both content-aware and distortion-aware comprehension. Specifically, the JPEG codec compresses images by dividing them into $8 \times 8$ blocks, necessitating networks to possess a robust capability for aggregating local information in order to effectively restore compression artifacts. Although the transformer block of PromptIR~\cite{promptir} has computational advantages, it fails to meet the local information extraction needs of blind CIR due to its reliance on global-wise token attention. Recently, the shifted window-based attention mechanism has demonstrated great potential in IR tasks~\cite{liang2021swinir,hat,li2023cswin2sr,conde2022swin2sr}, primarily because it integrates the ability to extract local features with the capacity for long-range pixel dependency modeling enabled by window-shifting. Notably, HAT~\cite{hat} integrates a channel attention block in parallel with the original shifted window-based attention block, as convolution facilitates the transformer to enhance representation abilities~\cite{li2201uniformer,wu2021cvt,yuan2021incorporating,xiao2021early}. Furthermore, HAT introduces the overlapped window-based attention to directly establish cross-window connections, boosting the representative capabilities. Concretely, HAT establishes a residual hybrid attention group (RHAG) by combining these two designs, \textit{possessing strong local and global information extraction abilities that meet our needs for blind CIR}. Therefore, we replace the first two transformer blocks in both the downsampling and upsampling stages with RHAG. The structure of RHAG is illustrated in Fig.~\ref{fig:RHAG}

\noindent \textbf{Discussion.} As image features go deeper, \ie, stage $L_3$ and $L_4$, local feature extraction becomes less important that global information modeling~\cite{resnet,vaswani2017attention}. Besides, the spatial resolution of feature maps is insufficient for window-shifting in deeper stages. Therefore, we only apply RHAG to the first two stages.

\begin{table*}[h!]
\centering
\caption{Quantitative comparison for blind CIR. Notice that, for a fair comparison on the scale of the training dataset, we train PromptIR\dag and PromptCIR with only DF2K, marking with *. Results are tested in terms of  PSNR$\uparrow$/SSIM$\uparrow$/PSNRB$\uparrow$. The best performances are \textbf{bolded}.}
\resizebox{\textwidth}{!}{
\begin{tabular}{c|ccccc}
\hline
 & JPEG & FBCNN~\cite{fbcnn} & Wang \etal~\cite{blindcontrastive} & PromptIR\dag*~\cite{promptir} & PromptCIR*  \\ \hline
Training data & - & DF2K & DIV2K+BSDS500(train) & DF2K & DF2K  \\ \hline
Params(M) & - & 71.92 & 11.11 & 36.54 & 34.75  \\ \hline
LIVE1 &  30.02/0.872/28.18 & 32.08/0.904/31.44 & 32.26/0.907/31.70 & 32.34/0.907/31.75 & \textbf{32.45/0.908/31.87}  \\
BSDS500 & 30.58/0.876/28.29 & 32.41/0.903/31.45 & \textbf{32.75/0.908/31.78} & 32.67/0.907/31.67 & 32.74/0.907/31.76 \\
ICB & 33.87/0.842/33.05 & 35.90/0.865/36.33 & 35.93/0.865/36.35 & 36.35/0.868/36.75 & \textbf{36.41/0.870/36.79}\\
DIV2K & 31.71/0.885/30.02 & 33.96/0.918/33.41 & 34.13/0.919/33.62 & 34.38/\textbf{0.922}/33.81 & \textbf{34.45/0.922/33.88} \\ \hline
 
\end{tabular}}
\label{tab:blind}
\end{table*}

\begin{table*}[h]
\centering
\caption{Quantitative comparison for non-blind CIR evaluation. Notice that, for a fair comparison on the scale of the training dataset, we train PromptIR\dag and PromptCIR with only DF2K, marking with *. Results are tested in terms of  PSNR$\uparrow$/SSIM$\uparrow$/PSNRB$\uparrow$. The best and second best performances are \textbf{bolded} and \underline{underlined}, respectively. We also include PromptCIR trained with DF2K and LSDIR for a comprehensive comparison.}
\resizebox{\textwidth}{!}{
\begin{tabular}{c|c|ccccc|c}
\hline
Dataset & QF & JPEG & FBCNN~\cite{fbcnn} & Wang \etal~\cite{blindcontrastive} & PromptIR\dag*~\cite{promptir} & PromptCIR* & PromptCIR \\ \hline
\multirow{4}{*}{LIVE1} & 10 & 25.69/0.743/24.20 &  27.77/0.803/27.51 & 27.80/0.805/27.57 & 27.98/0.806/27.66 & \underline{28.00/0.806/27.73} & \textbf{28.28/0.813/27.94}\\
 & 20 & 28.06/0.826/26.49 &  30.11/0.868/29.70 & 30.23/0.872/29.85 & 30.31/0.872/29.82 & \underline{30.32/0.872/29.86} & \textbf{30.61/0.876/30.12} \\
 & 30 & 29.37/0.861/27.84 &  31.43/0.897/30.92 & 31.58/0.900/\underline{31.13} & 31.66/0.900/31.07 & \underline{31.69/0.901}/31.10 & \textbf{31.96/0.904/31.38}\\
 & 40 & 30.28/0.882/28.84 &  32.34/0.913/31.80 & 32.53/0.916/32.04 & 32.58/0.915/31.96 & \underline{32.64/0.917/32.12} & \textbf{32.88/0.919/32.26} \\ \hline
 \multirow{4}{*}{ICB} & 10 & 29.44/0.757/28.53 & 32.18/0.815/32.15 & 32.05/0.813/32.04 & 32.70/0.821/33.01 & \underline{32.73/0.822/33.12} & \textbf{33.15/0.825/33.45} \\
 & 20 & 32.01/0.806/31.11 & 34.38/0.844/34.34 & 34.32/0.842/34.31
 & 34.84/0.847/35.21 & \underline{34.89/0.848/35.24} & \textbf{35.21/0.850/35.57} \\
 & 30 & 33.20/0.831/32.35 & 35.41/0.857/35.35 & 35.37/0.856/35.35 & 35.85/0.860/36.24 & \underline{35.89/0.861/36.26} & \textbf{36.19/0.863/36.58} \\
 & 40 & 33.95/0.840/33.14 & 36.02/0.866/35.95 & 35.99/0.860/35.97 & 36.45/0.869/36.85 & \underline{36.52/0.871/36.92} & \textbf{36.78/0.872/37.18}\\ \hline
\end{tabular}
}
\label{tab:non-blind}
\end{table*}

\subsection{Overall Pipeline}
To tackle the challenging blind CIR problem, we propose a prompt-learning-based framework, named PromptCIR. Compared with previous blind CIR and prompt-learning-based methods, PromptCIR has several advantages: 1) We use lightweight dynamic prompts~\cite{li2024ucip} to implicitly encode content-aware and distortion-aware information as flexible guidance for restoration networks. Prompts have more spatial adaptability than numerical predicted quality factors. 2) We use RHAG~\cite{hat} in the first two stages to enhance the local and global modeling capabilities of network, which is more helpful for removing compression artifacts.

The framework of PromptCIR is demonstrated in Fig.~\ref{fig:framework}. Given a compressed input image $\text{X}_\text{{Input}} \in \mathbb{R}^{H \times W \times 3}$, where H and W are the spatial resolutions, the network first uses a convolution layer to convert image into feature $F_0 \in \mathbb{R}^{H \times W \times C}$, where C is the embedded channel number. $F_0$ further passes through a 4-stage U-shape encoder-decoder structure. In the encoder stage, the image feature is iteratively downsampled to $F_l \in \mathbb{R}^{\frac{H}{8} \times \frac{W}{8} \times 8C}$. In the decoder stage, the image feature will gradually upsampled from $F_l$ to the original spatial resolution. We utilize prompts to implicitly learn the content-aware and distortion-aware information as a guidance for each decoder stage. Additionally, following~\cite{zamir2022restormer,promptir}, a refinement block with 4 transformer blocks is used to further enhance image features. A convolution layer is leveraged to project the channel number of restored feature $F_r$ from $C$ back to 3. Finally, we get restored clean image $\text{X}_\text{{Restored}} \in \mathbb{R}^{H \times W \times 3}$ by adding restored feature $F_r$ with $\text{X}_\text{{Input}}$.

\begin{figure*}[ht]
    \vspace{-1mm}
    \centering
    \includegraphics[width=1.0\linewidth]{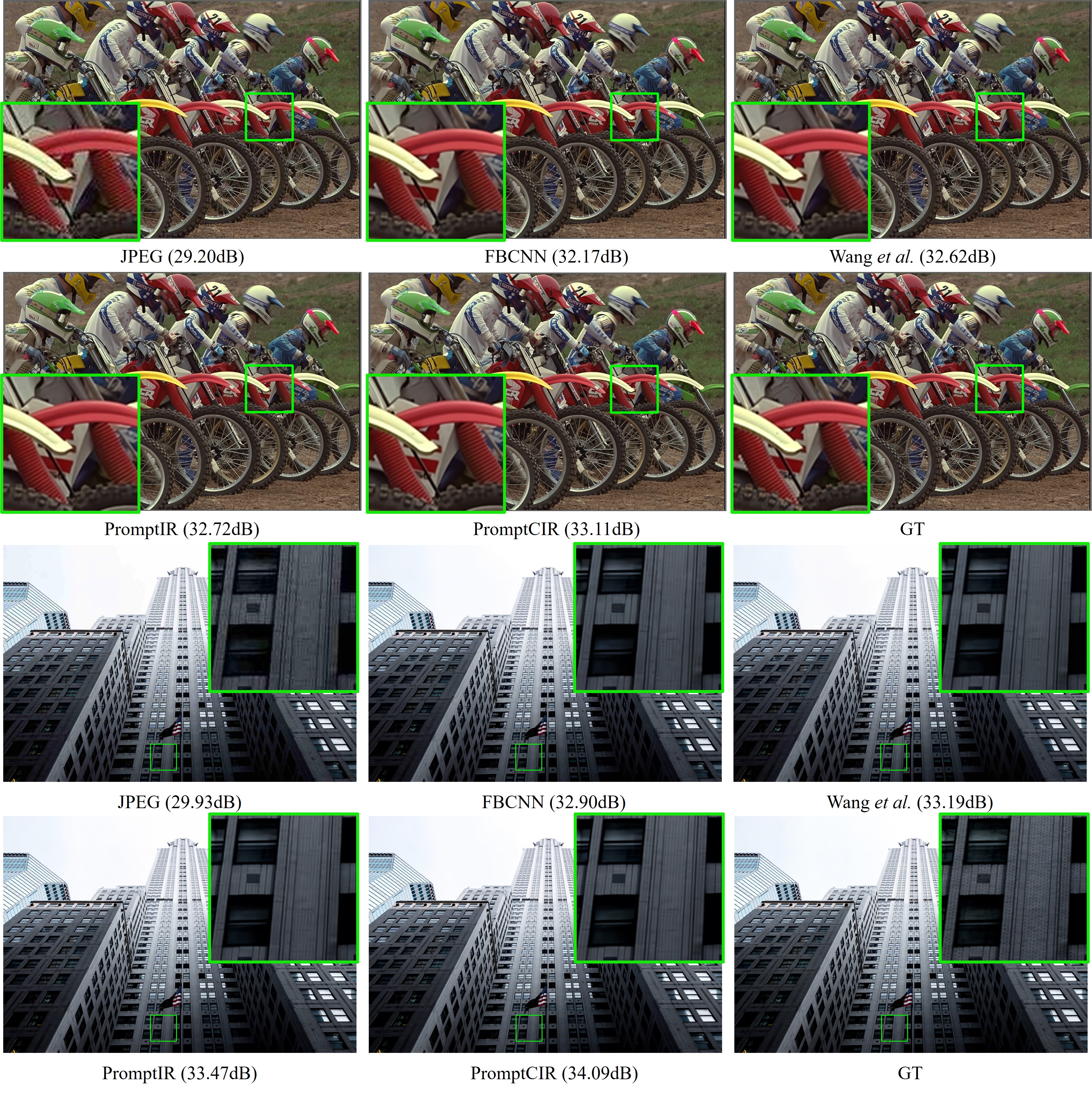}
    \caption{Qualitative comparisons between different methods on blind CIR. Zoom in for better views. (Upper: LIVE1\_bikes. Bottom: DIV2K\_0846)}
    \label{fig:blind}
    \vspace{-4.5mm}
\end{figure*}

\section{Experiments}
\subsection{Implementation Details}
\noindent \textbf{Training Datasets.} We use DF2K (800 images from DIV2K~\cite{DIV2K} and 2650 images from Flickr2K~\cite{Flickr2K}) and LSDIR~\cite{li2023lsdir} to form our training datasets. It is noteworthy that LSDIR is a recently proposed large-scale high-quality training datasets for IR tasks. Following the suggestion of NTIRE 2024 official code\footnote{https://codalab.lisn.upsaclay.fr/competitions/17548}, we apply PIL toolkit to generate compressed training images at 7 quality factors ranging from 10 to 70 with a step size of 10.

\noindent \textbf{Optimization Details.} We train our PromptCIR by 8 NVIDIA Tesla V100 16GB GPUs with a total batchsize of 24. To explore the upper limits of our model, we adopt a two-stage training strategy to optimize our PromptCIR. Specifically, in the first stage, the model is pre-trained on 7 aforementioned quality factors, fully developing its content-aware information extraction abilities. In the second stage, the model is fine-tuned on online compressed images with quality factors randomly selected from [10, 70]. At this stage, the model focuses more on distortion-aware information encoding, while still retaining the ability to extract content-aware information through prompt bases~\cite{li2024ucip}.

During the whole training process, the training images are paired-cropped into $128\times 128$ patches following~\cite{fbcnn,promptir}, while applying random flips and rotations. In the first stage, We adopt AdamW~\cite{adamw} optimizer to optimize our model with an initial learning rate of 2e-4. A CosineAnnealing scheduler is employed to decay the learning rate to 1e-6, with the total number of iterations set to 800k. In the second stage, we apply the same strategy with modification of the initial learning rate to 1e-4 and the iterations to 600k.

\noindent \textbf{Network parameters.} We follow the same settings of block numbers ([4,6,6,8]) and head numbers ([1,2,4,8]) of PromptIR~\cite{promptir}. The only difference is that, in the first two stages of the downsampling and upsampling stages, the block number refers to the number of successive HABs, and the head number refers to the count of heads used in the shifted window-based attention blocks.

\begin{figure*}
    \centering
    \includegraphics[width=1.0\linewidth]{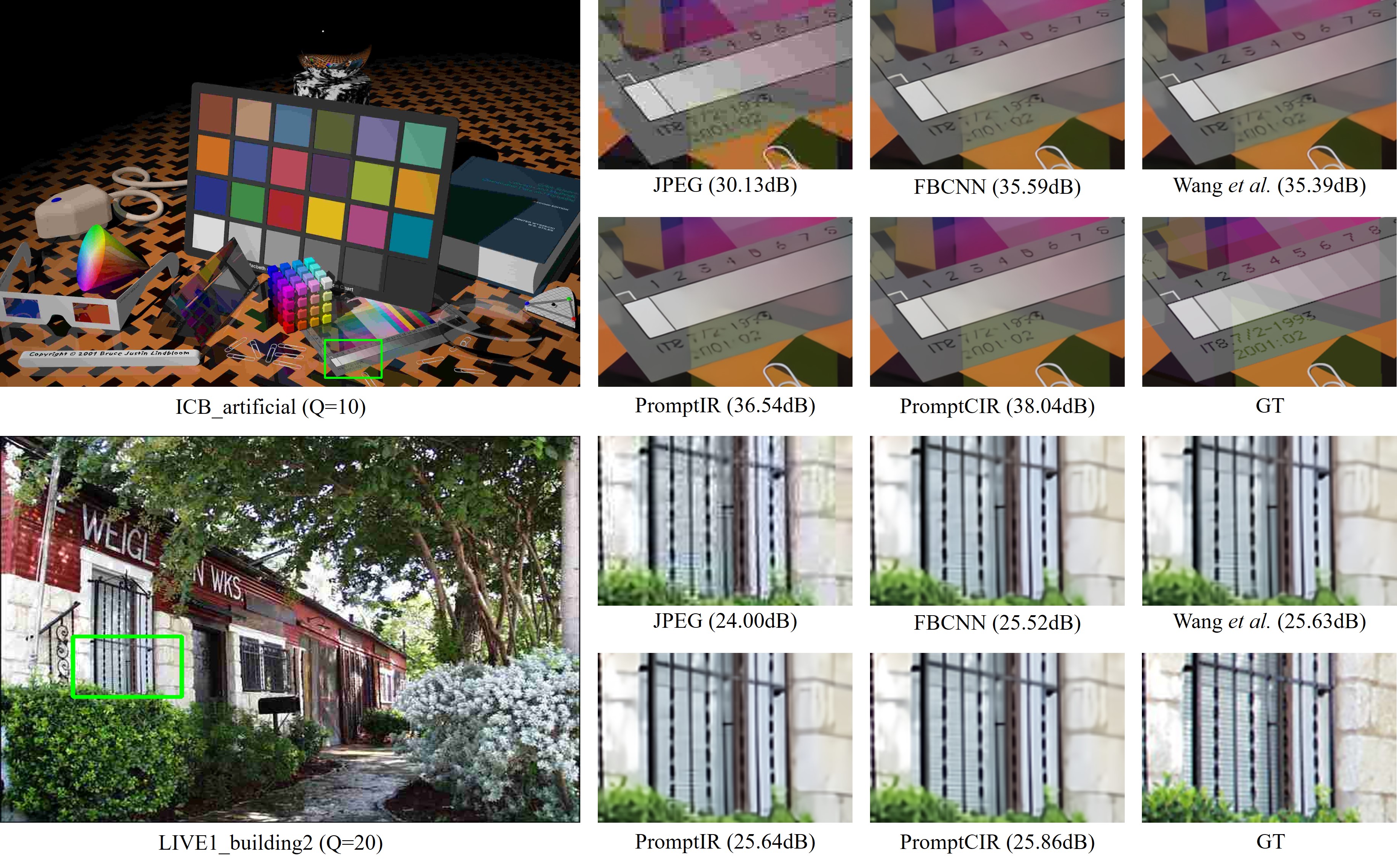}
    \caption{Qualitative comparisons between different methods on ICB with quality factor of 10 and LIVE1 with quality factor of 20. Our PromptCIR achieves visual pleasant restoration results. Zoom in for better views.}
    \label{fig:non-blind}
\end{figure*}

\subsection{Blind Compressed Image Restoration}
We evaluate the performance of PromptCIR against the contrastive learning-based method~\cite{blindcontrastive} (denoted as Wang \etal), FBCNN~\cite{fbcnn}, and the prompt-learning-based method PromptIR~\cite{promptir}. Notice that, as the released PromptIR are not trained on CIR, we retrain PromptIR with the same strategy as PromptCIR to ensure a fair comparison, denoted as PromptIR\dag. We include three compression benchmarks LIVE1~\cite{live1}, BSDS500~\cite{bsds} and ICB~\cite{ICB}. The high-quality images are randomly compressed by a quality factor ranging from 10 to 70, following the NTIRE 2024~\cite{yang2024ntire} implementation. Furthermore, we adopt the NTIRE 2024 official-released blind DIV2K validation set for evaluation. The results are tested on the RGB channel with PSNR, SSIM, and PSNRB. It is noteworthy that, for a fair comparison on the scale of training datasets, we also train PromptIR\dag and our PromptCIR with only DF2K, denoted as PromptIR\dag* and PromptCIR*.

As demonstrated in Tab.~\ref{tab:blind}, we have two observations: 1) prompt-learning-based methods have significant advantages on high-resolution datasets, \ie, ICB~\cite{ICB} and DIV2K~\cite{DIV2K}. We attribute the reason to the strong representation ability of prompts in modeling content-aware information, which surpasses that of numerical quality factor prediction and contrastive learning. 2) Benefiting from window-based attention, our PromptCIR achieves better performance than PromptIR, which indicates the importance of extracting local information. For qualitative results shown in Fig.~\ref{fig:blind}, we can observe that PromptCIR is capable of restoring texture details without blurring or blocking effects.

\subsection{Non-blind CIR Evaluation}
Non-blind CIR targets for evaluating model performance at a specific compression level, while the model is still \textit{trained in a blind way}. It serves as an important validation baseline for blind methods. Following previous works~\cite{QGAC,blindcontrastive,fbcnn}, we evaluate our PromptCIR on LIVE1~\cite{live1} and ICB~\cite{ICB} with quality factors [10,20,30,40]. The results are listed in Tab.~\ref{tab:non-blind}. We can observed that, apart from PromptCIR, PromptCIR* achieves the highest performance on almost all quality factors and datasets. As anticipated, owing to the utilization of a large dataset for training, PromptCIR has realized an additional average performance PSNR gain of about 0.3dB compared to PromptCIR*. This performance gain is inseparable from the content-aware and distortion-aware adaptability of the dynamic prompts~\cite{li2024ucip}, as well as the robust representation capabilities of the RHAG~\cite{hat} module. The subjective results are shown in Fig.~\ref{fig:non-blind}. PromptCIR restores characters and building textures more clearly and vividly, while achieving an average 0.86dB performance gain compared to PromptIR.

\subsection{Ablation Studies}

\begin{figure}
    \centering
    \includegraphics[width=1.0\linewidth]{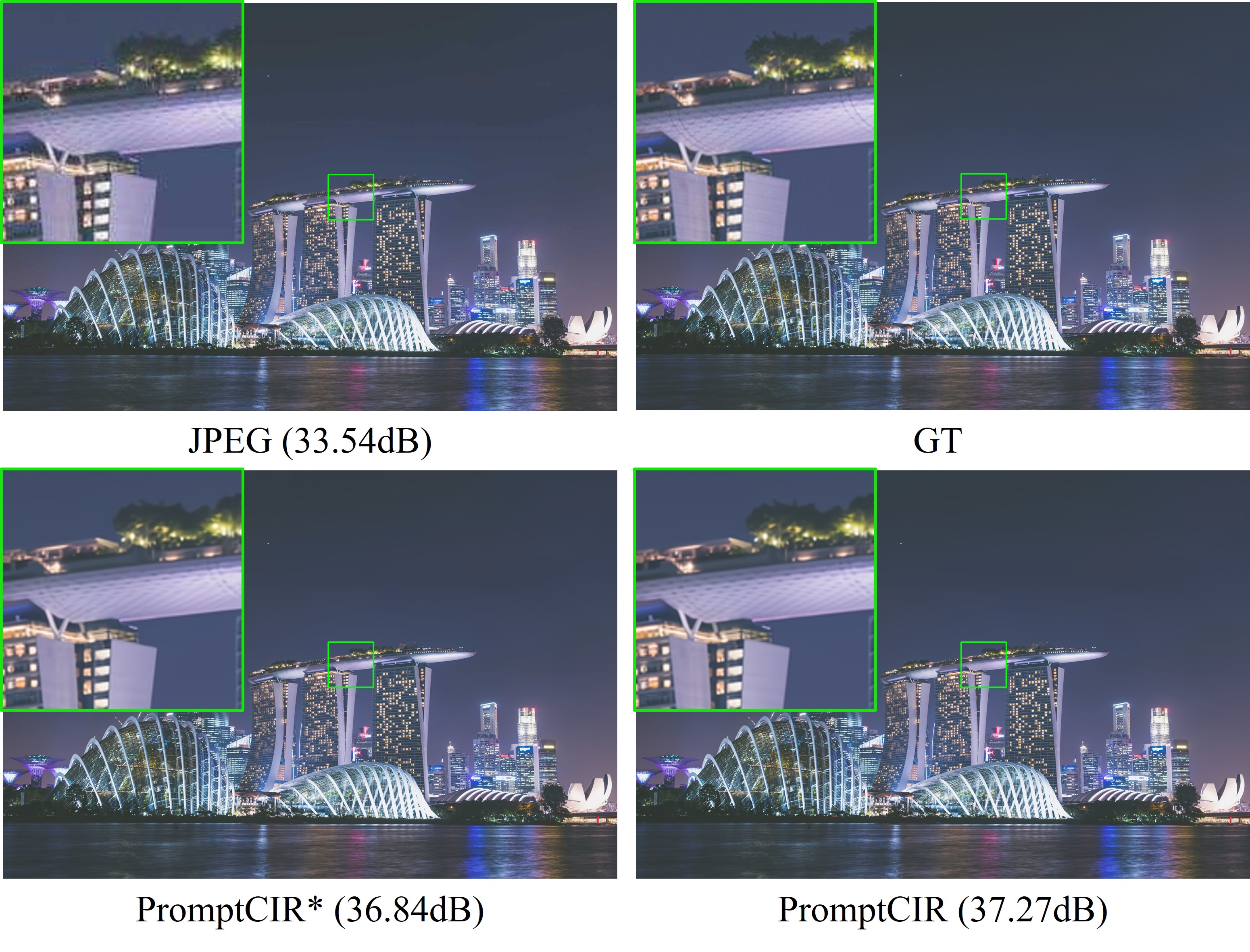}
    \caption{Qualitative comparisons between different training dataset scales. Zoom in for better views. (DIV2K\_0814)}
    \label{fig:ab_dataset}
\end{figure}

\noindent \textbf{The Effectiveness of Large-scale Training Datasets.} In IR tasks, models are usually optimized in supervised ways~\cite{edsr,dncnn,li2020multi,zamir2022restormer,liang2021swinir}. Typically, these models learn the pixel-wise or distribution-wise~\cite{wang2018esrganRRDB,realesrgan,bsrgan,xia2023diffir,lin2023diffbir,stablesr} relationships between distorted and clean image pairs through training datasets. However, in real-world scenarios, such relationships are complex and diverse. Merely relying on a few thousand training images is insufficient for the model to learn the correspondence between distorted and clean images applicable to real-world scenarios~\cite{li2023lsdir}. Therefore, a larger training dataset has become crucial needs in the field of IR. Luckily, recent work LSDIR~\cite{li2023lsdir} has introduced a large-scale training dataset consisting of 84991 high-quality images, which is a hundred times larger than popular DIV2K~\cite{DIV2K} training dataset, significantly boost the performance of IR networks. Inspired by this, we explore the performance of PromptCIR by optimizing it with LSDIR. The quantitative comparisons are given in Tab.~\ref{tab:dataset}. As demonstrated, large-scale training datasets confer two significant advantages: 1) It is noteworthy that with increased resolution and diversity of the dataset (\eg, ICB with $5k \times 3k$ resolution), we observe a larger performance gain. This can be attributed to the fact that larger-scale training datasets offer more rich texture details, thereby enhancing the abilities of models to establish complex mappings between compressed and clean images. 2) As demonstrated in Sec.~\ref{sec:hab}, we employ RHAG~\cite{hat} for its enhanced representation capabilities in extracting both local and global information. It naturally follows that RHAG stands to achieve more performance gain than the transposed transformer block~\cite{zamir2022restormer,promptir} from larger-scale training datasets. We provide qualitative comparisons in Fig.~\ref{fig:ab_dataset}. As illustrated, PromptCIR trained with LSDIR achieves better texture restoration qualities. 

\begin{table}[]
\centering
\caption{Quantitative comparison for different scales of training datasets. Results are tested in terms of  PSNR$\uparrow$/SSIM$\uparrow$/PSNRB$\uparrow$ on blind validation sets. We highlight the performance gain of PSNR by $\uparrow_\textbf{value}$ for more intuitive comparisons.}
\resizebox{\linewidth}{!}{
\begin{tabular}{l|c|cc}
\hline
Dataset & Method & DF2K & DF2K+LSDIR \\ \hline
\multirow{2}{*}{LIVE1} & PromptIR & 32.34/0.907/31.75 & $32.42_{\uparrow\textbf{0.08}}$/0.908/31.83 \\
 & Ours & 32.45/0.908/31.87 & $32.65_{\uparrow\textbf{0.11}}$/0.910/32.06 \\ \hline
\multirow{2}{*}{BSDS500} & PromptIR & 32.67/0.907/31.67 & $32.76_{\uparrow\textbf{0.09}}$/0.908/31.72 \\
 & Ours & 32.74/0.907/31.76 & $32.92_{\uparrow\textbf{0.18}}$/0.910/31.83 \\ \hline
\multirow{2}{*}{ICB} & PromptIR & 36.35/0.868/36.75 & $36.54_{\uparrow\textbf{0.19}}$/0.870/36.93 \\
 & Ours & 36.41/0.870/36.79 & $36.70_{\uparrow\textbf{0.29}}$/0.781/37.09 \\ \hline
\multirow{2}{*}{DIV2K} & PromptIR & 34.38/0.922/33.81 & $34.52_{\uparrow\textbf{0.14}}$/0.923/33.90 \\
 & Ours & 34.45/0.922/33.88 & $34.68_{\uparrow\textbf{0.23}}$/0.925/34.08 \\ \hline
\end{tabular}
\label{tab:dataset}
}
\end{table}

\noindent \textbf{The effectiveness of RHAG and DPM.}
As demonstrated in Sec.~\ref{sec:dpm} and Sec.~\ref{sec:hab}, blind CIR requires network to possess powerful extraction abilities for both local and global content-aware and distortion-aware information. To explore the effectiveness of RHAG~\cite{hat} and DPM~\cite{li2024ucip}, we conduct ablation studies in Tab.~\ref{tab:rhag}. Based on the quantitative results, we have validated that: 1) Compared to the original prompt design of PromptIR~\cite{promptir}, DPM~\cite{li2024ucip} achieves better performance with dynamic prompt bases. 2) Compared to the original transformer block~\cite{zamir2022restormer}, RHAG enjoys stronger representation abilities for CIR task.

\begin{table}[]
\centering
\caption{Quantitative comparison for RHAG and DPM. Results are tested in terms of  PSNR$\uparrow$/SSIM$\uparrow$ on blind DIV2K set. We adopt \checkmark to indicate the usage of specific modules.}
\resizebox{0.8\linewidth}{!}{
\begin{tabular}{l|cc|c}
\hline
Method & RHAG & DPM & PSNR/SSIM \\ \hline
PromptIR &  &  & 34.52/0.923 \\ \hline
\multirow{3}{*}{PromptCIR} &  & \checkmark & 34.56/0.924 \\
 & \checkmark &  & 34.62/0.924 \\
 & \checkmark & \checkmark & \textbf{34.68/0.925} \\ \hline
\end{tabular}
}
\label{tab:rhag}
\end{table}

\section{Conclusion}
In this paper, we present PromptCIR, the prompt-learning-based CIR model, which adopts lightweight prompts as effective content-aware and distortion-aware information guidance for blind CIR task. Different from existing prediction-based networks that estimate numerical quality factors, our PromptCIR has advantages of spatial-wise adaptabilities with dynamic prompt bases. Moreover, we explore the effectiveness of large-scale training dataset, which is capable of further boosting the qualities of restored images. Our method has achieved first place in the NTIRE 2024 challenge of blind compressed image enhancement track. Extensive experiments on blind and non-blind benchmarks have demonstrated the superiority of proposed PromptCIR.

{
    \small
    \bibliographystyle{ieeenat_fullname}
    \bibliography{main}
}


\end{document}